\documentclass[sigconf]{acmart}

\AtBeginDocument{%
  }

\copyrightyear{2022}
\acmYear{2022}
\setcopyright{acmcopyright}\acmConference[ICGSP 2022]{2022 The 6th International Conference on Graphics and Signal Processing (ICGSP)}{July 1--3, 2022}{Chiba, Japan}
\acmBooktitle{2022 The 6th International Conference on Graphics and Signal Processing (ICGSP) (ICGSP 2022), July 1--3, 2022, Chiba, Japan}
\acmPrice{15.00}
\acmDOI{10.1145/3561518.3561519}
\acmISBN{978-1-4503-9637-0/22/06}

\begin{document}

\title{Face Emotion Recognization Using Dataset Augmentation Based on Neural Network}


\author{Mengyu Rao}

\affiliation{%
  \institution{Fuzhou University}
}

\author{Ruyi Bao}
\affiliation{%
  \institution{University of Nottingham}
}
\author{Liangshun Dong*}
\affiliation{
 \institution{Shanghai Jiao Tong University}
}


\begin{abstract}
  Face expression plays a critical role during the daily life, and people cannot live without face emotion. With the development of technology, many methods of facial expression recognition have been proposed. However, from traditional methods to deep learning methods, few of them pay attention to the hybrid data augmentation, which can help improve the robustness of models. Therefore, a method of hybrid data augmentation is highlighted in this paper. The hybrid data augmentation is a method of combining several effective data augmentation. In the experiments, the technique is applied on four basic networks and the results are compared to the baseline models. After applying this technique, the results show that four benchmark models have higher performance than those previously. This approach is simple and robust in terms of data augmentation, which makes it applied in the real world in the future. Besides the results show versatility of the technique as all of our experiments get better results.
\end{abstract}



\maketitle

\section{Introduction}
Facial expression is one of the most external indications of a person's feelings and emotions. In daily conversation, according to the psychologist, only 7\% and 38\% of information is communicated through words and sounds respective, while up to 55\% is through facial expression \cite{Mehrabian1967InferenceOA}. It plays an important role in coordinating interpersonal relationships.
Ekman and Friesen \cite{Ekman1971ConstantsAC} recognized six essential emotions in the nineteenth century depending on a cross-cultural study  \cite{4}, which indicated that people feel each basic emotion in the same fashion despite culture.
As a branch of the field of analyzing sentiment \cite{GAN2020104827}, facial expression recognition offers broad application prospects in a variety of domains, including the interaction between humans and computers \cite{8606936}, healthcare \cite{Ilyas2018FacialER}, and behavior monitoring \cite{Rabhi2018ARE}. Therefore, many researchers have devoted themselves to facial expression recognition.
In this paper, an effective hybrid data augmentation method is used. This approach is operated on two public datasets, and four benchmark models see some remarkable results.

\section{RELATED WORKS}

\subsection{VggNet}
The VGG model \cite{Simonyan15} was posted by the Visual Geometry Group team at Oxford University. The primary goal of this architecture is to demonstrate how the its final performance can be impacted by increasing network depth. In VGG, 7×7 convolution kernels are replaced by three 3×3 convolution kernels, and 5×5 convolution kernels are replaced by two 3×3 convolution kernels. The main goal of the change is to make sure that the depth of the network and the impact of the neural network can be ameliorated with the condition of the same perceptual field.
%
%

\subsection{ResNet}
The ResNet \cite{666826} model won first place in the ImageNet competition \cite{networks} held in 2015. The problem that deepening the model can decrease the accuracy was solved by this work. Due to the proposed residual block, it is easy to learn the identity mapping, even though stacked. If there are numerous blocks, redundant blocks can also learn the identity mapping with the help of the residual block. Furthermore, it improves the effectiveness of SGD optimization, which can optimize the network in deeper. What is more, no additional parameters and computational complexity are introduced. Only a very simple addition operation is performed and the complexity is negligible compared to the convolution operation. The ResNet architecture is shown in Figure 1.
\begin{figure}[h]
	\centering
	\includegraphics[width=\linewidth]{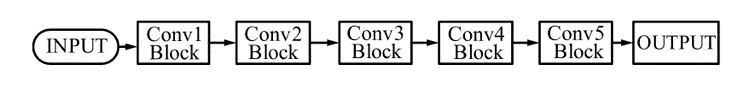}
	\caption{The structure of ResNet}
\end{figure}
%
%
%
%
\subsection{Xception}
The Xception \cite{8099678} model is an upgraded version of the InceptionV3 \cite{7780677} model. Chollet F offers a new structure of deep convolutional neural network named Xception that replaces the Inception module with a depthwise separable convolution. The residual network and the depthwise separable convolution are the fundamental components of this network. Xception is typically composed of 36 convolutional layers grouped into 14 blocks, with 12 blocks in the middle containing all linear residual connections. Simultaneously, the model holds the properties of depthwise separable convolution\cite{7780459} since the model executes spatial layer-by-layer convolution on every channel of the inputs individually, and then conducts point-by-point convolution on the output.
\begin{figure}[h]
	\centering
	\includegraphics[width=\linewidth]{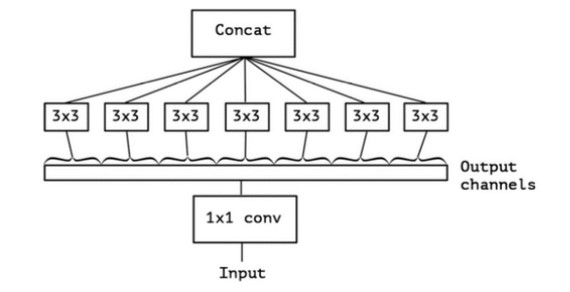}
	\caption{The network structure of Xception}
\end{figure}
\section{APPROACH}
\subsection{HDA: hybrid data augmentation}
\subsubsection{Horizontal Flip.}
Geometric transformation is one of the most basic methods to augment data. Because of the particularity of the images, which means that facial expression images emoticons do not undergo a large degree of distortion and rotation in most cases, the horizontal flip (HF) is used to ensure that these images are consistent. Every image in the original dataset is horizontally flipped to create a mirror image. The formula of this method is shown in Eq. (1).\\
$$
\left[
\begin{array}{l}
	x^{'}\\
	y^{'}
\end{array}
\right]
=
\left[
\begin{array}{l}
	width-1\\
	0
\end{array}
\right]
+
\left[
\begin{array}{ll}
	-1 & 0\\
	0 & 1
\end{array}
\right]\cdot
\left[
\begin{array}{l}
	x\\
	y
\end{array}
\right]
$$
\subsubsection{Gaussian Noise.}

Gaussian noise (GN) represents a kind of statistical noise that has a probability density function equivalent to that of the normal distribution. In the proposed approach, the training images are added with Gaussian noise to simulate noise that may happen in the real world so that the model can become robust against the original images. The formula of Gaussian noise can be written as Eq. (2).
$$
GN(x,y)=\frac{1}{2\pi\sigma_{1}\sigma_{2}\sqrt{1-\rho^{2}}}e^{(-\frac{1}{2(1-\rho^{2})})(\frac{(x-\mu_{1})^{2}}{\sigma_{1}^{2}}-\frac{2\rho(x-\mu_{1})(y-\mu_{2})}{\sigma_{1}\sigma_{2}}+\frac{(y-\mu_{2})^{2}}{\sigma_{2}^{2}})}
$$
\section{EXPERIMENTS AND RESULTS}
\subsection{Datasets}
Following related work on facial emotion recognition, these experiments are conducted on the two benchmark public face emotion datasets: Ck+ dataset\cite{Lucey2010TheEC}, and Fer2013  dataset \cite{GOODFELLOW201559}.
Ck+: The most widely utilized laboratory-controlled dataset for facial expression recognition is the Extended Cohn–Kanade (Ck+) Dataset\cite{Lucey2010TheEC} (some samples are shown in Figure 3). Sequences that change from neutral to peak expression are included in the Ck+ dataset. Extraction of the final 1 to 3 frames which have peak formation and the first frame of every sequence is the most common data selection approach for evaluation. Then, people are divided into n groups for person-independent n-fold cross-validation experiments, where n is typically between 5, 8, and 10.
\begin{figure}[h]
	\centering
	\includegraphics[width=\linewidth]{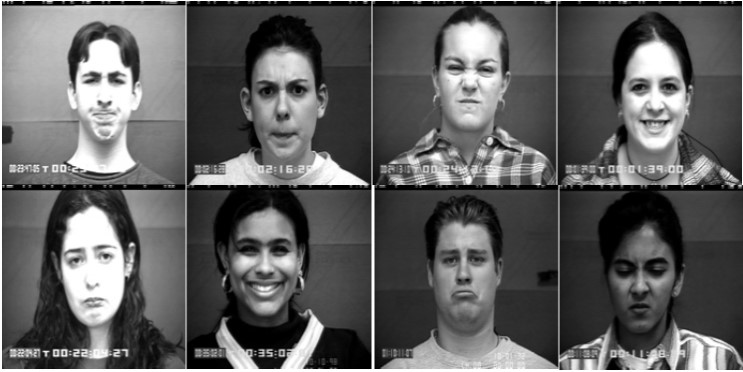}
	\caption{Some samples of the Ck+ dataset}
\end{figure}
Fer2013: Fer2013 \cite{GOODFELLOW201559} is a large-scale dataset acquired automatically by the Google image search API (some samples are shown in Figure 4). 35887 images are contained in the Fer2013 dataset, and each image is labeled as one of the seven basic emotions. All of the images in this dataset are grayscale images. Furthermore, this dataset contains 547 disgusted images, 5121 fear images, 4953 angry images, 8989 happy images, 4002 surprised images, 6077 sad images, and 6198 neutral images.

\begin{figure}[h]
	\centering
	\includegraphics[width=\linewidth]{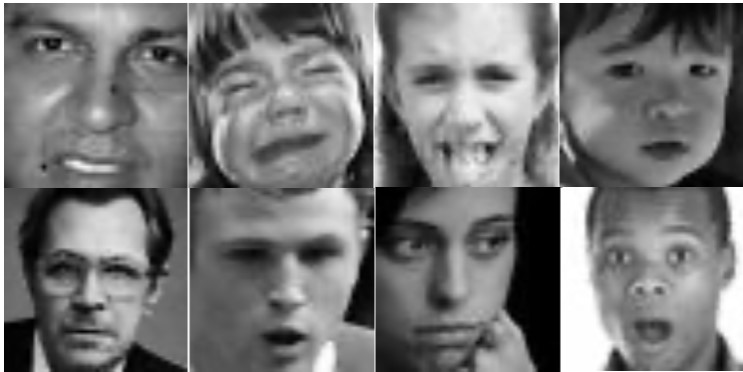}
	\caption{Some samples of the Fer2013 dataset}
\end{figure}
Some information about the datasets are shown in Table 1.
\begin{table}[H]
	
	\begin{center}
		\begin{tabular}{|c|c|c|c|}
			\hline
			\textbf{Dataest} & \textbf{Number}& \textbf{Number of Emotion} &\textbf{Gender} \\
			\hline
			JAFFE & 213 images+ & 7 & Female \\
			\hline
			CK+ & 593 videos & 7 &Female \& Male \\
			\hline
			Fer2013 & 35886 images & 7  &Female \& Male \\
			\hline
			
		\end{tabular}
		\label{tab2}
	\end{center}
\caption{Datasets information}
\end{table}

\subsection{Experimental settings}
The experiments are implemented via PyTorch \cite{NEURIPS2019_bdbca288}, and the NVIDIA GTX 2080Ti with 4 CPU cores and 13 Gigabytes of RAM is used for experiments. The Adam optimizer \cite{Kingma2015AdamAM} is used to train the networks with a learning rate of 2e-4 and betas of 0.9 and 0.999. The best model is selected through the principle of selection of the best accuracy of several experiments.

\subsection{Experimental evaluation metric}
Face emotion recognition can be viewed as a multi-classification problem. Accuracy (Acc) is used as the evaluation metric in this paper, and the calculation formula is as follows:
Acc=(TP+TN)/(TP+TN+FP+FN)	  	(3)	
In this formula, TP stands for a positive sample predicted by the model as a positive sample, TN stands for a negative sample predicted by the model as a negative sample, FP stands for a negative sample predicted by the model as a positive sample, and FN stands for a positive sample predicted by the model as a negative sample.

\subsection{Performance on Ck+ dataset}
This dataset consists of 8 emotions with a total of 981 trainable images, all of which are 640 pixels × 490 pixels in size. 593 trainable images from 7 expressions are selected for the experiment (shown in Figure 5). However, the background of the volunteers in the pictures is larger than the face images. Therefore, the image size of the dataset is processed to 48×48 pixels, which is convenient for the model input size to be uniform.
\begin{figure}[h]
	\centering
	\includegraphics[width=\linewidth]{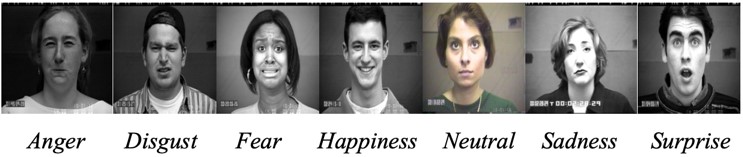}
	\caption{Images of different expressions in the Ck+ dataset}
\end{figure} 
The results of the Ck+ dataset and the HDACK+ dataset are shown in Table 2, where both the Ck+ dataset and the HDACK+ dataset are divided in the ratio of 8:1:1 for training, validating, and testing respectively. The batch size used is 32, and 20 epochs are used for training.
\begin{table}[H]
	
	\begin{center}
		\begin{tabular}{|c|c|c|}
			\hline
			\textbf{Model} & \textbf{Ck+}& \textbf{HDACK+} \\
			\hline
			Vgg19 & 74.19\% & 97.80\% \\
			\hline
			Resnet18 & 90.32\% & 100\% \\
			\hline
			Resnet50 & 95.70\% & 99.73\% \\
			\hline
			Xception & 83.87\% & 99.73\% \\
			\hline
		\end{tabular}
		\label{tab2}
	\end{center}
\caption{Accuracy Comparisons of Different Models on Ck+ dataset}
\end{table}
\begin{figure}[h]
	\centering
	\includegraphics[width=\linewidth]{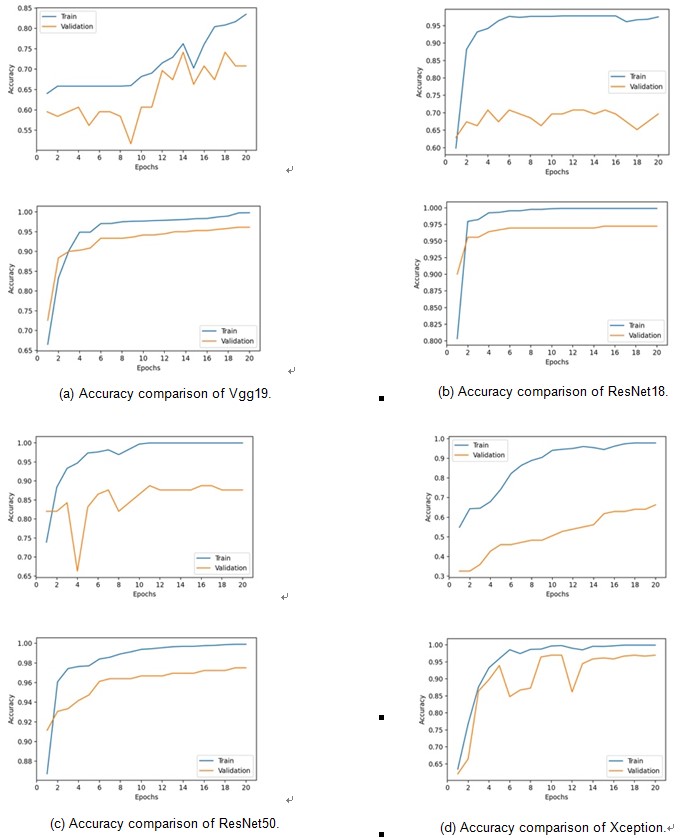}
	\caption{Accuracy comparison of different models on Ck+ dataset and HDACK+ dataset. (Best viewed in color)}
\end{figure} 
Tables 2 and Figure 6 indicate the accuracies of the Ck+ dataset and those of the HDACK+ dataset. The dataset with data augmentation always has a higher performance in most models, and the ResNet18 model achieved 100\% testing accuracy in the HDACK+ dataset. After analyzing the results, significant improvement in accuracy can be seen on all models, especially for Vgg19. The accuracy comparison of Vgg19 shows that the accuracy improves by 23.61\%. The ResNet50 network is the least improved one, but the accuracy is also 4.03\% higher than the original data.
Furthermore, the figures show that the data augmentation improves the feature learning for every model. For the left one of each column, the Ck+ dataset is used. The training accuracies have an upward tendency, but the validation accuracies always have some shocks, even increasing slowly. Especially for ResNet18, the curve always fluctuates around 70\%. However, when comes to the right one of each column, the HDACK+ dataset is used. Both the curve of training accuracy and the curve of validation have an increasing toward, and the validation accuracies of all figures in the right position come to a peak of around 98\%.

\subsection{Performance on Fer2013 dataset}

After rejecting wrongly labeled frames, each image was registered and scaled to 48×48 pixels. There are 28,709, 3,589 and 3,589 images respectively for training, validation, and  testing with seven expression labels in the Fer2013 dataset (shown in Figure 7).
\begin{figure}[h]
	\centering
	\includegraphics[width=\linewidth]{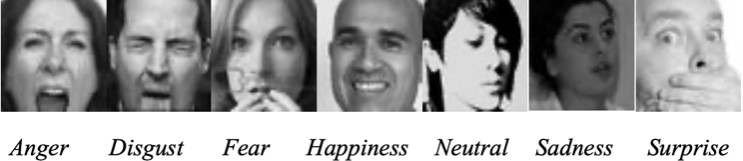}
	\caption{Images of different expressions in the Fer2013 dataset}
\end{figure} 
The results of the Fer2013 dataset and the HDAFer2013 dataset are presented in Table 3, where both the Fer2013 dataset and the HDAFer2013 dataset are divided in the ratio of 8:1:1 for training, validating, and testing respectively. The batch size used is 32, and 20 epochs are used for training.

\begin{table}[H]
	
	\begin{center}
		\begin{tabular}{|c|c|c|}
			\hline
			\textbf{Model} & \textbf{Fer2013}& $\mathbf{Fer2013_{DA}}$ \\
			\hline
			Vgg19 & 62.13\% & 84.87\% \\
			\hline
			Resnet18 & 65.67\% & 88.32\% \\
			\hline
			Resnet50 & 62.55\% & 88.17\% \\
			\hline
			Xception & 0.6035\% & 82.68\% \\
			\hline
		\end{tabular}
		\label{tab2}
	\end{center}
\caption{Accuracy Comparisons of Different Models on Fer2013 dataset}
\end{table}

\begin{figure}[h]
	\centering
	\includegraphics[width=\linewidth]{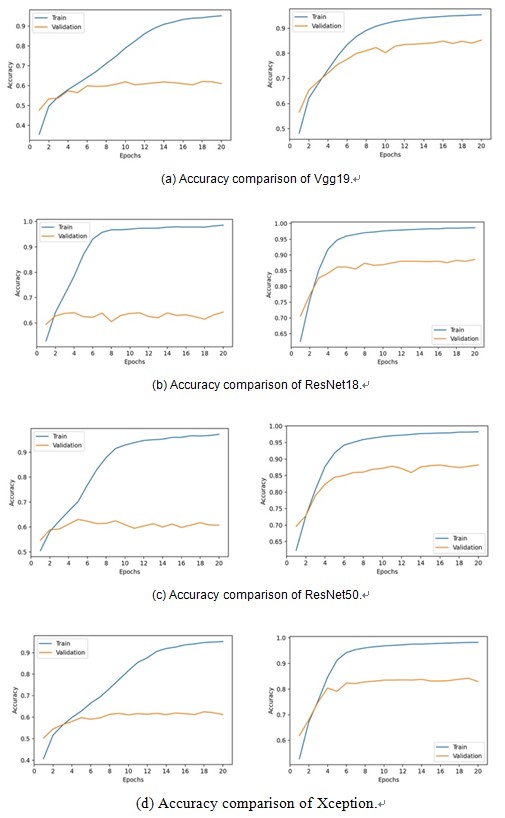}
	\caption{Accuracy comparison of different models on Fer2013 dataset and HDAFer2013 dataset. (Best viewed in color)}
\end{figure} 
Tables 3 and Figure 8 indicate the accuracies of the Fer2013 dataset and those of the HDAFer2013 dataset. The dataset used data augmentation always has a higher performance in most models. The results show that improvement in accuracy can be seen in all models. The accuracy comparison of ResNet50 shows that the accuracy improves by 25.62\%. The Xception network is the least improved, but the accuracy on it is also 22.33\% higher than that on the original data.
Furthermore, the figures show that the data augmentation improves the feature learning for every model. For the left one of each column, the Fer2013 dataset is used. The training accuracies have an upward tendency, but the validation accuracies always fluctuate around 63\%. However, when comes to the right one of each column, the HDAFer2013 dataset is used. Both the curve of training accuracy and the curve of validation have an increasing toward, and the validation accuracies of all figures in the right position come to a peak above 82\%.

\section{CONCLUSION}
In this study, a hybrid data augmentation method of the dataset, which has a high performance in some models, is presented. Because convolutional neural networks require more samples for training to get accurate and robust result, the hybrid data augmentation method is used to enlarge the number of samples. After applying the technique, the numbers of images in both the Ck+ dataset and the Fer2013 dataset have increased, and four benchmark models have higher performance than those previously. This approach is simple and robust in terms of data augmentation, which makes it applicable in the real world in the future.

\bibliographystyle{ACM-Reference-Format}
\bibliography{bib}


\begin{thebibliography}{17}


\ifx \showCODEN    \undefined \def \showCODEN     #1{\unskip}     \fi
\ifx \showDOI      \undefined \def \showDOI       #1{#1}\fi
\ifx \showISBNx    \undefined \def \showISBNx     #1{\unskip}     \fi
\ifx \showISBNxiii \undefined \def \showISBNxiii  #1{\unskip}     \fi
\ifx \showISSN     \undefined \def \showISSN      #1{\unskip}     \fi
\ifx \showLCCN     \undefined \def \showLCCN      #1{\unskip}     \fi
\ifx \shownote     \undefined \def \shownote      #1{#1}          \fi
\ifx \showarticletitle \undefined \def \showarticletitle #1{#1}   \fi
\ifx \showURL      \undefined \def \showURL       {\relax}        \fi
\providecommand\bibfield[2]{#2}
\providecommand\bibinfo[2]{#2}
\providecommand\natexlab[1]{#1}
\providecommand\showeprint[2][]{arXiv:#2}

\bibitem[\protect\citeauthoryear{??}{net}{[n.d.]}]%
        {networks}
 \bibinfo{year}{[n.d.]}\natexlab{}.
\newblock \bibinfo{booktitle}{\emph{ImageNet Large Scale Visual Recognition
  Challenge (ILSVRC)}}.
\newblock
\urldef\tempurl%
\url{https://www.image-net.org/challenges/LSVRC/}
\showURL{%
\tempurl}
\newblock
\shownote{(2022, Jul 10).}


\bibitem[\protect\citeauthoryear{Carreira, Madeira, and Silva}{Carreira
  et~al\mbox{.}}{1998}]%
        {666826}
\bibfield{author}{\bibinfo{person}{J. Carreira}, \bibinfo{person}{H. Madeira},
  {and} \bibinfo{person}{J.G. Silva}.} \bibinfo{year}{1998}\natexlab{}.
\newblock \showarticletitle{Xception: a technique for the experimental
  evaluation of dependability in modern computers}.
\newblock \bibinfo{journal}{\emph{IEEE Transactions on Software Engineering}}
  \bibinfo{volume}{24}, \bibinfo{number}{2} (\bibinfo{year}{1998}),
  \bibinfo{pages}{125--136}.
\newblock
\urldef\tempurl%
\url{https://doi.org/10.1109/32.666826}
\showDOI{\tempurl}


\bibitem[\protect\citeauthoryear{Chollet}{Chollet}{2017}]%
        {8099678}
\bibfield{author}{\bibinfo{person}{François Chollet}.}
  \bibinfo{year}{2017}\natexlab{}.
\newblock \showarticletitle{Xception: Deep Learning with Depthwise Separable
  Convolutions}. In \bibinfo{booktitle}{\emph{2017 IEEE Conference on Computer
  Vision and Pattern Recognition (CVPR)}}. \bibinfo{pages}{1800--1807}.
\newblock
\urldef\tempurl%
\url{https://doi.org/10.1109/CVPR.2017.195}
\showDOI{\tempurl}


\bibitem[\protect\citeauthoryear{Deng, Pang, Zhang, Pang, Yang, and Yang}{Deng
  et~al\mbox{.}}{2019}]%
        {8606936}
\bibfield{author}{\bibinfo{person}{Jia Deng}, \bibinfo{person}{Gaoyang Pang},
  \bibinfo{person}{Zhiyu Zhang}, \bibinfo{person}{Zhibo Pang},
  \bibinfo{person}{Huayong Yang}, {and} \bibinfo{person}{Geng Yang}.}
  \bibinfo{year}{2019}\natexlab{}.
\newblock \showarticletitle{cGAN Based Facial Expression Recognition for
  Human-Robot Interaction}.
\newblock \bibinfo{journal}{\emph{IEEE Access}}  \bibinfo{volume}{7}
  (\bibinfo{year}{2019}), \bibinfo{pages}{9848--9859}.
\newblock
\urldef\tempurl%
\url{https://doi.org/10.1109/ACCESS.2019.2891668}
\showDOI{\tempurl}


\bibitem[\protect\citeauthoryear{Ekman}{Ekman}{1994}]%
        {4}
\bibfield{author}{\bibinfo{person}{Pual Ekman}.}
  \bibinfo{year}{1994}\natexlab{}.
\newblock \showarticletitle{Strong evidence for universals in facial
  expressions: A reply to Russell's mistaken critique.}
\newblock \bibinfo{journal}{\emph{Psychological Bulletin}}
  \bibinfo{volume}{115 2} (\bibinfo{year}{1994}), \bibinfo{pages}{268--287}.
\newblock


\bibitem[\protect\citeauthoryear{Ekman and Friesen}{Ekman and Friesen}{1971}]%
        {Ekman1971ConstantsAC}
\bibfield{author}{\bibinfo{person}{Paul Ekman} {and} \bibinfo{person}{W~V
  Friesen}.} \bibinfo{year}{1971}\natexlab{}.
\newblock \showarticletitle{Constants across cultures in the face and emotion.}
\newblock \bibinfo{journal}{\emph{Journal of personality and social
  psychology}}  \bibinfo{volume}{17 2} (\bibinfo{year}{1971}),
  \bibinfo{pages}{124--9}.
\newblock


\bibitem[\protect\citeauthoryear{Gan, Wang, Zhang, and Wang}{Gan
  et~al\mbox{.}}{2020}]%
        {GAN2020104827}
\bibfield{author}{\bibinfo{person}{Chenquan Gan}, \bibinfo{person}{Lu Wang},
  \bibinfo{person}{Zufan Zhang}, {and} \bibinfo{person}{Zhangyi Wang}.}
  \bibinfo{year}{2020}\natexlab{}.
\newblock \showarticletitle{Sparse attention based separable dilated
  convolutional neural network for targeted sentiment analysis}.
\newblock \bibinfo{journal}{\emph{Knowledge-Based Systems}}
  \bibinfo{volume}{188} (\bibinfo{year}{2020}), \bibinfo{pages}{104827}.
\newblock
\showISSN{0950-7051}
\urldef\tempurl%
\url{https://doi.org/10.1016/j.knosys.2019.06.035}
\showDOI{\tempurl}


\bibitem[\protect\citeauthoryear{Goodfellow, Erhan, {Luc Carrier}, Courville,
  Mirza, Hamner, Cukierski, Tang, Thaler, Lee, Zhou, Ramaiah, Feng, Li, Wang,
  Athanasakis, Shawe-Taylor, Milakov, Park, Ionescu, Popescu, Grozea, Bergstra,
  Xie, Romaszko, Xu, Chuang, and Bengio}{Goodfellow et~al\mbox{.}}{2015}]%
        {GOODFELLOW201559}
\bibfield{author}{\bibinfo{person}{Ian~J. Goodfellow}, \bibinfo{person}{Dumitru
  Erhan}, \bibinfo{person}{Pierre {Luc Carrier}}, \bibinfo{person}{Aaron
  Courville}, \bibinfo{person}{Mehdi Mirza}, \bibinfo{person}{Ben Hamner},
  \bibinfo{person}{Will Cukierski}, \bibinfo{person}{Yichuan Tang},
  \bibinfo{person}{David Thaler}, \bibinfo{person}{Dong-Hyun Lee},
  \bibinfo{person}{Yingbo Zhou}, \bibinfo{person}{Chetan Ramaiah},
  \bibinfo{person}{Fangxiang Feng}, \bibinfo{person}{Ruifan Li},
  \bibinfo{person}{Xiaojie Wang}, \bibinfo{person}{Dimitris Athanasakis},
  \bibinfo{person}{John Shawe-Taylor}, \bibinfo{person}{Maxim Milakov},
  \bibinfo{person}{John Park}, \bibinfo{person}{Radu Ionescu},
  \bibinfo{person}{Marius Popescu}, \bibinfo{person}{Cristian Grozea},
  \bibinfo{person}{James Bergstra}, \bibinfo{person}{Jingjing Xie},
  \bibinfo{person}{Lukasz Romaszko}, \bibinfo{person}{Bing Xu},
  \bibinfo{person}{Zhang Chuang}, {and} \bibinfo{person}{Yoshua Bengio}.}
  \bibinfo{year}{2015}\natexlab{}.
\newblock \showarticletitle{Challenges in representation learning: A report on
  three machine learning contests}.
\newblock \bibinfo{journal}{\emph{Neural Networks}}  \bibinfo{volume}{64}
  (\bibinfo{year}{2015}), \bibinfo{pages}{59--63}.
\newblock
\showISSN{0893-6080}
\urldef\tempurl%
\url{https://doi.org/10.1016/j.neunet.2014.09.005}
\showDOI{\tempurl}
\newblock
\shownote{Special Issue on “Deep Learning of Representations”.}


\bibitem[\protect\citeauthoryear{He, Zhang, Ren, and Sun}{He
  et~al\mbox{.}}{2016}]%
        {7780459}
\bibfield{author}{\bibinfo{person}{Kaiming He}, \bibinfo{person}{Xiangyu
  Zhang}, \bibinfo{person}{Shaoqing Ren}, {and} \bibinfo{person}{Jian Sun}.}
  \bibinfo{year}{2016}\natexlab{}.
\newblock \showarticletitle{Deep Residual Learning for Image Recognition}. In
  \bibinfo{booktitle}{\emph{2016 IEEE Conference on Computer Vision and Pattern
  Recognition (CVPR)}}. \bibinfo{pages}{770--778}.
\newblock
\urldef\tempurl%
\url{https://doi.org/10.1109/CVPR.2016.90}
\showDOI{\tempurl}


\bibitem[\protect\citeauthoryear{Ilyas, Haque, Rehm, Nasrollahi, and
  Moeslund}{Ilyas et~al\mbox{.}}{2018}]%
        {Ilyas2018FacialER}
\bibfield{author}{\bibinfo{person}{Chaudhary Muhammad~Aqdus Ilyas},
  \bibinfo{person}{Mohammad~Ahsanul Haque}, \bibinfo{person}{Matthias Rehm},
  \bibinfo{person}{Kamal Nasrollahi}, {and} \bibinfo{person}{Thomas~Baltzer
  Moeslund}.} \bibinfo{year}{2018}\natexlab{}.
\newblock \showarticletitle{Facial Expression Recognition for Traumatic Brain
  Injured Patients}. In \bibinfo{booktitle}{\emph{VISIGRAPP}}.
\newblock


\bibitem[\protect\citeauthoryear{Kingma and Ba}{Kingma and Ba}{2015}]%
        {Kingma2015AdamAM}
\bibfield{author}{\bibinfo{person}{Diederik~P. Kingma} {and}
  \bibinfo{person}{Jimmy Ba}.} \bibinfo{year}{2015}\natexlab{}.
\newblock \showarticletitle{Adam: A Method for Stochastic Optimization}.
\newblock \bibinfo{journal}{\emph{CoRR}}  \bibinfo{volume}{abs/1412.6980}
  (\bibinfo{year}{2015}).
\newblock


\bibitem[\protect\citeauthoryear{Lucey, Cohn, Kanade, Saragih, Ambadar, and
  Matthews}{Lucey et~al\mbox{.}}{2010}]%
        {Lucey2010TheEC}
\bibfield{author}{\bibinfo{person}{Patrick Lucey}, \bibinfo{person}{Jeffrey~F.
  Cohn}, \bibinfo{person}{Takeo Kanade}, \bibinfo{person}{Jason~M. Saragih},
  \bibinfo{person}{Zara Ambadar}, {and} \bibinfo{person}{I. Matthews}.}
  \bibinfo{year}{2010}\natexlab{}.
\newblock \showarticletitle{The Extended Cohn-Kanade Dataset (CK+): A complete
  dataset for action unit and emotion-specified expression}.
\newblock \bibinfo{journal}{\emph{2010 IEEE Computer Society Conference on
  Computer Vision and Pattern Recognition - Workshops}} (\bibinfo{year}{2010}),
  \bibinfo{pages}{94--101}.
\newblock


\bibitem[\protect\citeauthoryear{Mehrabian and Ferris}{Mehrabian and
  Ferris}{1967}]%
        {Mehrabian1967InferenceOA}
\bibfield{author}{\bibinfo{person}{Albert Mehrabian} {and}
  \bibinfo{person}{Suzanne Ferris}.} \bibinfo{year}{1967}\natexlab{}.
\newblock \showarticletitle{Inference of attitudes from nonverbal communication
  in two channels.}
\newblock \bibinfo{journal}{\emph{Journal of consulting psychology}}
  \bibinfo{volume}{31 3} (\bibinfo{year}{1967}), \bibinfo{pages}{248--52}.
\newblock


\bibitem[\protect\citeauthoryear{Paszke, Gross, Massa, Lerer, Bradbury, Chanan,
  Killeen, Lin, Gimelshein, Antiga, Desmaison, Kopf, Yang, DeVito, Raison,
  Tejani, Chilamkurthy, Steiner, Fang, Bai, and Chintala}{Paszke
  et~al\mbox{.}}{2019}]%
        {NEURIPS2019_bdbca288}
\bibfield{author}{\bibinfo{person}{Adam Paszke}, \bibinfo{person}{Sam Gross},
  \bibinfo{person}{Francisco Massa}, \bibinfo{person}{Adam Lerer},
  \bibinfo{person}{James Bradbury}, \bibinfo{person}{Gregory Chanan},
  \bibinfo{person}{Trevor Killeen}, \bibinfo{person}{Zeming Lin},
  \bibinfo{person}{Natalia Gimelshein}, \bibinfo{person}{Luca Antiga},
  \bibinfo{person}{Alban Desmaison}, \bibinfo{person}{Andreas Kopf},
  \bibinfo{person}{Edward Yang}, \bibinfo{person}{Zachary DeVito},
  \bibinfo{person}{Martin Raison}, \bibinfo{person}{Alykhan Tejani},
  \bibinfo{person}{Sasank Chilamkurthy}, \bibinfo{person}{Benoit Steiner},
  \bibinfo{person}{Lu Fang}, \bibinfo{person}{Junjie Bai}, {and}
  \bibinfo{person}{Soumith Chintala}.} \bibinfo{year}{2019}\natexlab{}.
\newblock \showarticletitle{PyTorch: An Imperative Style, High-Performance Deep
  Learning Library}. In \bibinfo{booktitle}{\emph{Advances in Neural
  Information Processing Systems}},
  \bibfield{editor}{\bibinfo{person}{H.~Wallach},
  \bibinfo{person}{H.~Larochelle}, \bibinfo{person}{A.~Beygelzimer},
  \bibinfo{person}{F.~d\textquotesingle Alch\'{e}-Buc},
  \bibinfo{person}{E.~Fox}, {and} \bibinfo{person}{R.~Garnett}} (Eds.),
  Vol.~\bibinfo{volume}{32}. \bibinfo{publisher}{Curran Associates, Inc.}
\newblock
\urldef\tempurl%
\url{https://proceedings.neurips.cc/paper/2019/file/bdbca288fee7f92f2bfa9f7012727740-Paper.pdf}
\showURL{%
\tempurl}


\bibitem[\protect\citeauthoryear{Rabhi, Mrabet, Fnaiech, and Sayadi}{Rabhi
  et~al\mbox{.}}{2018}]%
        {Rabhi2018ARE}
\bibfield{author}{\bibinfo{person}{Yassine Rabhi}, \bibinfo{person}{Makrem
  Mrabet}, \bibinfo{person}{Farhat Fnaiech}, {and} \bibinfo{person}{Mounir
  Sayadi}.} \bibinfo{year}{2018}\natexlab{}.
\newblock \showarticletitle{A real-time emotion recognition system for disabled
  persons}.
\newblock \bibinfo{journal}{\emph{2018 4th International Conference on Advanced
  Technologies for Signal and Image Processing (ATSIP)}}
  (\bibinfo{year}{2018}), \bibinfo{pages}{1--6}.
\newblock


\bibitem[\protect\citeauthoryear{Simonyan and Zisserman}{Simonyan and
  Zisserman}{2015}]%
        {Simonyan15}
\bibfield{author}{\bibinfo{person}{Karen Simonyan} {and}
  \bibinfo{person}{Andrew Zisserman}.} \bibinfo{year}{2015}\natexlab{}.
\newblock \showarticletitle{Very Deep Convolutional Networks for Large-Scale
  Image Recognition}. In \bibinfo{booktitle}{\emph{International Conference on
  Learning Representations}}.
\newblock


\bibitem[\protect\citeauthoryear{Szegedy, Vanhoucke, Ioffe, Shlens, and
  Wojna}{Szegedy et~al\mbox{.}}{2016}]%
        {7780677}
\bibfield{author}{\bibinfo{person}{Christian Szegedy}, \bibinfo{person}{Vincent
  Vanhoucke}, \bibinfo{person}{Sergey Ioffe}, \bibinfo{person}{Jon Shlens},
  {and} \bibinfo{person}{Zbigniew Wojna}.} \bibinfo{year}{2016}\natexlab{}.
\newblock \showarticletitle{Rethinking the Inception Architecture for Computer
  Vision}. In \bibinfo{booktitle}{\emph{2016 IEEE Conference on Computer Vision
  and Pattern Recognition (CVPR)}}. \bibinfo{pages}{2818--2826}.
\newblock
\urldef\tempurl%
\url{https://doi.org/10.1109/CVPR.2016.308}
\showDOI{\tempurl}


\end{thebibliography}
%
%
%
%
%
%
%
%
%
%

\end{document}